\documentclass[10pt, conference, compsocconf]{IEEEtran}

\usepackage{caption}
\usepackage{subcaption}

\usepackage[l2tabu,orthodox]{nag}

\usepackage[
backend=bibtex8, 
style=ieee, 
sorting=none, 
natbib=true, 
doi=false, 
isbn=false, 
url=false, 
eprint=false, 
maxcitenames=1, 
mincitenames=1,
maxbibnames=50
]{biblatex}

\usepackage[pdftex,colorlinks]{hyperref}

\usepackage[printonlyused]{acronym}

\usepackage{siunitx}
\usepackage[all]{nowidow}

\usepackage[dvipsnames]{xcolor}

\definecolor{tangelo}{rgb}{0.98, 0.3, 0.0}

\usepackage{lipsum}

\usepackage[pdftex]{graphicx}

\usepackage[normalem]{ulem} 

\usepackage{epstopdf}

\usepackage{import}

\graphicspath{{./latexGoodPractices/}}

\usepackage{booktabs}

\usepackage{tabu}

\usepackage{amssymb,amsfonts,amsmath,amscd}

\usepackage{bm} 

\newcommand{\set}[1]{\left\{#1\right\}}

\newcommand{\bbm}{\begin{bmatrix}}
\newcommand{\ebm}{\end{bmatrix}}

\acrodef{UAVs}{Unmanned Aerial Vehicles}
\acrodef{SubT}{DARPA Subterranean}
\acrodef{ICP}{Iterative Closest Point}
\acrodef{IMU}{Inertial Measurement Unit}
\acrodef{ESKF}{Error-State Kalman Filter}
\acrodef{POC}{Point Of Collision}
\acrodef{MEMS}{Micro Electro-Mechanical Systems}
\acrodef{FRMSD}{Fractional Root Mean Squared Distance}
\acrodef{EM}{Expectation-Maximization}
\acrodef{NM}{No Weighting}
\acrodef{TW}{Time-based Weighting}
\acrodef{VTW}{Velocity and Time-based Weighting}
\acrodef{GVTW}{Geometry, Velocity and Time-based Weighting}
\acrodef{SAW}{Scanning Angle-based Weighting}

\newcommand{\rot}[1]{\operatorname{rot}\!\left(#1\right)}
\newcommand{\T}[2]{\;\phantom{}_{#1}^{#2} \bm{T}}
\newcommand{\reading}[0]{\mathcal{P}}
\newcommand{\reference}[0]{\mathcal{Q}}
\newcommand{\scan}[0]{\mathcal{S}}
\newcommand{\point}[2][]{\,\phantom{}^{#1}\bm{p}_{#2}}
\newcommand{\cardinality}[0]{n}
\newcommand{\estlinvel}[1][]{\bm v_{#1}}
\newcommand{\estlinspeed}[1][]{v_{#1}}
\newcommand{\estangvel}[1][]{\bm \omega_{#1}}
\newcommand{\estangspeed}[1][]{\omega_{#1}}
\newcommand{\scale}[0]{\kappa}
\newcommand{\shape}[0]{\beta}
\newcommand{\linscale}[0]{\lambda}
\newcommand{\angscale}[0]{\phi}
\newcommand{\unlinspeed}[0]{\sigma_v}
\newcommand{\unangspeed}[0]{\sigma_\omega}
\newcommand{\normal}[1][]{\bm n_{#1}}
\newcommand{\distance}[1]{\operatorname{d}\!\left(#1\right)}
\newcommand{\transparam}[1]{\delta_{#1}}
\newcommand{\transparams}[0]{\bm{\transparam{}}}
\newcommand{\rotparam}[1]{\theta_{#1}}
\newcommand{\rotparams}[0]{\bm{\rotparam{}}}
\newcommand{\scanangle}[1][]{\gamma_{#1}}

\begin{document}

\makeatletter
\def\ps@IEEEtitlepagestyle{
	\def\@oddfoot{
		\begin{minipage}{\textwidth}
			\centering \scriptsize
			\copyright~2021 IEEE. Personal use of this material is permitted. Permission from IEEE must be obtained for all other uses, in any current or future media, including reprinting/republishing this material for advertising or promotional purposes, creating new collective works, for resale or redistribution to servers or lists, or reuse of any copyrighted component of this work in other works.
		\end{minipage}
	}
}
\makeatother
\clearpage
\thispagestyle{plain}
\pagestyle{plain}

%
\title{Lidar Scan Registration Robust to Extreme Motions}


\author{\IEEEauthorblockN{Simon-Pierre Desch\^{e}nes, Dominic Baril, Vladim\'{i}r Kubelka, Philippe Gigu\`{e}re, Fran\c{c}ois Pomerleau}
\IEEEauthorblockA{Northern Robotics Laboratory, Universit\'{e} Laval, Quebec City, Quebec, Canada\\
$\{$simon-pierre.deschenes, francois.pomerleau$\}$@norlab.ulaval.ca}
}


%


\maketitle

\begin{abstract}
Registration algorithms, such as \ac{ICP}, have proven effective in mobile robot localization algorithms over the last decades.
However, they are susceptible to failure when a robot sustains extreme velocities and accelerations.
For example, this kind of motion can happen after a collision, causing a point cloud to be heavily skewed.
While point cloud de-skewing methods have been explored in the past to increase localization and mapping accuracy, these methods still rely on highly accurate odometry systems or ideal navigation conditions.
In this paper, we present a method taking into account the remaining motion uncertainties of the trajectory used to de-skew a point cloud along with the environment geometry to increase the robustness of current registration algorithms.
We compare our method to three other solutions in a test bench producing 3D maps with peak accelerations of \SI[detect-weight=true]{200}{m/s^2} and \SI[detect-weight=true]{800}{rad/s^2}.
In these extreme scenarios, we demonstrate that our method decreases the error by \SI[detect-weight=true]{9.26}{\%} in translation and by \SI[detect-weight=true]{21.84}{\%} in rotation.
The proposed method is generic enough to be integrated to many variants of weighted \ac{ICP} without adaptation and supports localization robustness in harsher terrains.


\end{abstract}

\begin{IEEEkeywords}
lidar, point weighting, registration, ICP, localization, mapping, de-skewing, collision, extreme motion

\end{IEEEkeywords}

%
\IEEEpeerreviewmaketitle

\section{Introduction}
\acresetall

Time-critical operations, such as search and rescue, require mobile robots to be deployed in increasingly complex environments and at higher velocities.
In these conditions, the risk of such a system to sustain a collision is inevitably increased.
Thus, the ability to recover from collisions is fundamental to expand the autonomy of robots in large and unstructured environments.
This research topic has received a lot of attention over the last few years but was mostly related to \ac{UAVs} being mechanically collision-resilient \cite{Briod2014, Kornatowski2017, Dilaveroglu2020}. 
Also related to flying, highly reactive controllers have been explored to recover from collisions \cite{Dicker2017, Wang2020}.
However, to our knowledge, no work has been done on increasing the robustness of localization and mapping specifically to collisions.

\begin{figure}[htbp]
    \centering
    \includegraphics[width=0.95\columnwidth]{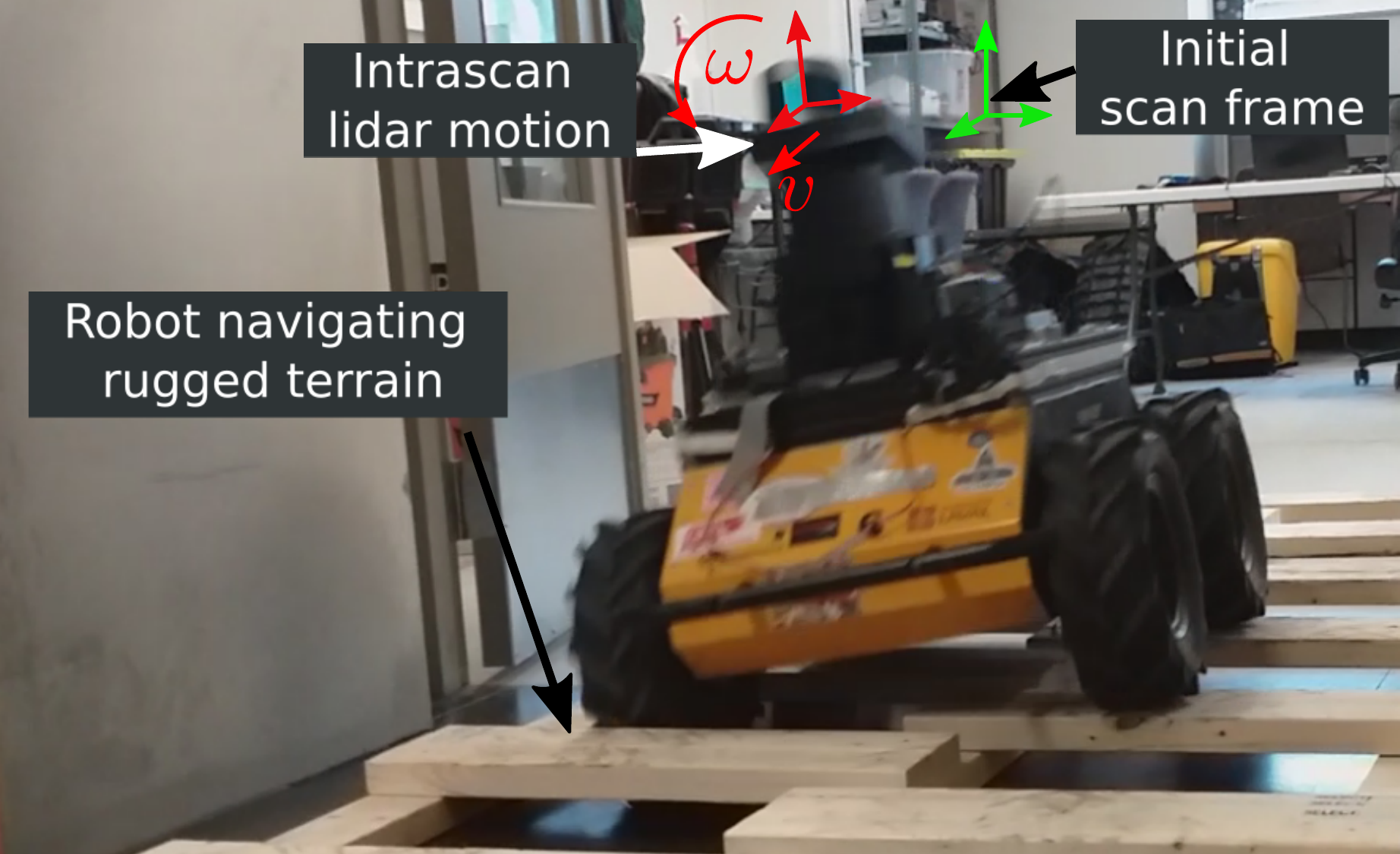}
    \caption{
    A mobile robot navigating on controlled rugged terrain.
    Due to the high velocities sustained by the robot, to which a lidar sensor is attached, rapid motion is induced within the acquisition time of a scan.
    This motion introduces a point cloud skewing effect, potentially leading to large localization errors.
    In green is the fixed reference frame of the scan and in red is the moving frame during a scan.
    }
    \label{fig:husky_ninja}
\end{figure}

High velocities and accelerations cause severe distortions in data acquired by lidar sensors.
This is due to the assumption that all points from the same scan are measured from the same origin, even if the lidar is subject to high motion during a scan.
Data distortions are manifested through a skewing effect in the measured point cloud, causing errors in point locations.
This skewing effect is proportional to the lidar speed during a scan, which is typically high when the sensor sustains a collision, as shown in~\autoref{fig:husky_ninja}.
If significant enough, the scan skewing can lead to localization failure due to inconsistencies of the environment representation.
A naive solution would be to ignore incoming scans when sustaining a collision, but this would cause discontinuities in the robot trajectory, also potentially leading to localization failure.
To mitigate scan skewing, point cloud de-skewing methods have proven effective~\cite{He2020}.
Point cloud de-skewing consists of applying a correction to the measured point cloud, taking lidar motion within a scan into account.
However, state-of-the-art methods assume accurate odometry~\cite{Renzler2020}, which is jeopardized during extreme motions.

In this work, we address this challenge by proposing a model to compute weights for each point in a scan based on motion uncertainties and environment geometry.
These weights are then directly used in the minimization process of the registration algorithm to increase localization accuracy and robustness.
%
The key contribution of this work is threefold:
\begin{enumerate}
    \item a novel motion uncertainty-based point weighting model, allowing the robustness of registration algorithms to be improved when subject to high velocities and accelerations, requiring neither high-fidelity pose estimation nor prior knowledge of the environment;
    \item experimental comparison of our proposed approach with state-of-the-art approaches on a dataset gathered at high velocities and accelerations; and
    \item a highlight of the impact of point cloud skewing on the accuracy of localization and mapping.
\end{enumerate}


	
	


\section{Related Work}
\label{sec:related_work}

Over the last few decades, the \ac{ICP} registration algorithm \cite{Chen1991,Besl1992} has been widely used in robotics for localization and mapping.
Since its introduction, many variants of \ac{ICP} were proposed to increase its robustness.
For instance, in the work of \citet{Phillips2007}, the \ac{FRMSD} measure was introduced to allow simultaneous minimization of alignment error and estimation of overlap ratio between two point clouds.
This allowed the registration of point clouds with varying overlap ratios, without having to manually tune this parameter each time the algorithm is run, making registration more robust.
\citet{Granger2002} proposed an \ac{ICP} variant based on the \ac{EM} algorithm, which registers point clouds at multiple scales to increase registration robustness.
However, these solutions work at the point cloud level and do not take into account external factors that can affect scan accuracy.
Therefore, when point clouds suffer from distortions due to rapid sensor motion, the accuracy of the algorithm is reduced.
In this work, we propose a weighting model that can be integrated with prior efforts on the robustness of variable overlaps without modification.

To solve the issue of distortions, point cloud de-skewing algorithms were introduced.
In the work of \citet{Bosse2009}, scan matching and lidar trajectory were optimized jointly to de-skew and register point clouds.
Their algorithm performed well at linear speeds up to \SI{6}{m/s}, but at low angular speeds.
Moreover, it was tested under unknown accelerations.
In a two-step approach, \citet{Zhang2014} first registered consecutive scans to estimate the lidar motion between them.
Then, they used this motion estimate to correct point cloud distortions, by assuming constant linear and angular velocities during a lidar scan.
While showing interesting results at speeds up to \SI{0.5}{m/s}, their algorithm was not tested at high speeds and accelerations.
In the work of \citet{Renzler2020}, a high-accuracy odometry system is used to compute the wheeled vehicle pose at a rate of \SI{50}{Hz}, enabling correction of motion distortion in lidar point clouds.
Their algorithm was tested at linear speeds up to \SI{25}{m/s} and under linear accelerations up to \SI{10}{m/s^2}, but at unknown angular speeds.
\citet{Zhao2019} developed a four-step odometry and mapping method to achieve real-time mapping on highways.
They estimated the vehicle pose by integrating \ac{IMU} measurements, allowing for point cloud de-skewing.
Their method was tested at linear speeds up to \SI{20}{m/s} and at angular speeds up to approximately \SI{1.05}{rad/s}, but at low accelerations, inherent to highway navigation.
For their part, \citet{Qin2020} developped an \ac{ESKF} to localize in real time.
Using \ac{IMU} and lidar odometry, they were able to de-skew point cloud features.
Their algorithm was tested at unknown speeds and accelerations.
The Ceres solver was used in the work of \citet{He2020}, to refine the \ac{IMU} trajectory during a scan, thereby minimizing point cloud deformations after de-skewing.
Their algorithm was tested at a linear speed of up to \SI{6.94}{m/s}, but at unknown angular speeds and accelerations.
Furthermore, to avoid distortion caused by high displacement during a scan, they discarded \SI{10}{\%} of the data.
Our experimental setup pushed the validation of our model far beyond de-skewing algorithm capabilities, with linear speeds and accelerations up to \SI{3.5}{m/s} and \SI{200}{m/s^2}, with rotational speeds and accelerations up to \SI{11}{rad/s} and \SI{800}{rad/s^2}.

As can be seen, the aforementioned work either lacked testing at high speeds and high accelerations, did not perform well or needed the use of a high-accuracy odometry system to work properly when subject to aggressive motion.
Instead, we propose a solution to mitigate the effect of such motion using a weighting approach similar to \citet{Al-Nuaimi2016}.
In their work, weights were assigned to each point of a scan, based on azimuth scanning angle and local surface curvature to scale the alignment error of matched points.
While showing significant improvements in simulation, their solution was tested at unknown speeds and accelerations on real-world data and showed little improvement.
However, in the method we propose, more elaborate weights will be used and compared with their model. 
We will demonstrate significant improvements on real lidar scans acquired at high speeds and under high accelerations.
Moreover, our solution can be used in conjunction with de-skewing approaches for further improvement of localization accuracy, allowing an increase in performance of the aforementioned methods.






\section{Theory}
\label{sec:theory}

Typical 3D lidars sense their environment by sweeping it with an array of laser beams rotating around their vertical axis.
In order to construct a point cloud from the data acquired during a scan, the lidar is assumed to be stationary throughout the scan. 
All points are consequently presumed to be located within the same coordinate frame.
However, this assumption does not hold most of the time for a mobile robot, as it is highly probable that the sensor will move within a scan.
Consequently, when all points of a scan are expressed in the same coordinate frame without consideration for this lidar displacement, the resulting point cloud is skewed.
This skewing effect increases proportionally with the linear and angular displacements of the lidar during a scan.


%
Skewed point clouds will degrade the accuracy of any localization and mapping framework, starting with the registration algorithm.
Given a reading point cloud $\reading$ and a reference point cloud $\reference$, the \ac{ICP} algorithm aims at finding the transformation $\T{}{}$ that minimizes the alignment error between $\reading$ and $\reference$.
In the context of localization and mapping, the reading $\reading$ is the latest point cloud acquired with the lidar and the reference $\reference$ is the map of the environment available after registering the last point cloud.
We can recover the pose of a robot over a trajectory by concatenating all of the rigid transformations $\T{}{}$ obtained from past registrations since the start of the trajectory.
Of course, this pose estimation will drift as small errors will be integrated throughout the complete trajectory, and skewed point clouds will precipitate this phenomenon.

\begin{figure}[htbp]
    \centering
    \includegraphics[trim=0 30 0 40, width=0.8\columnwidth]{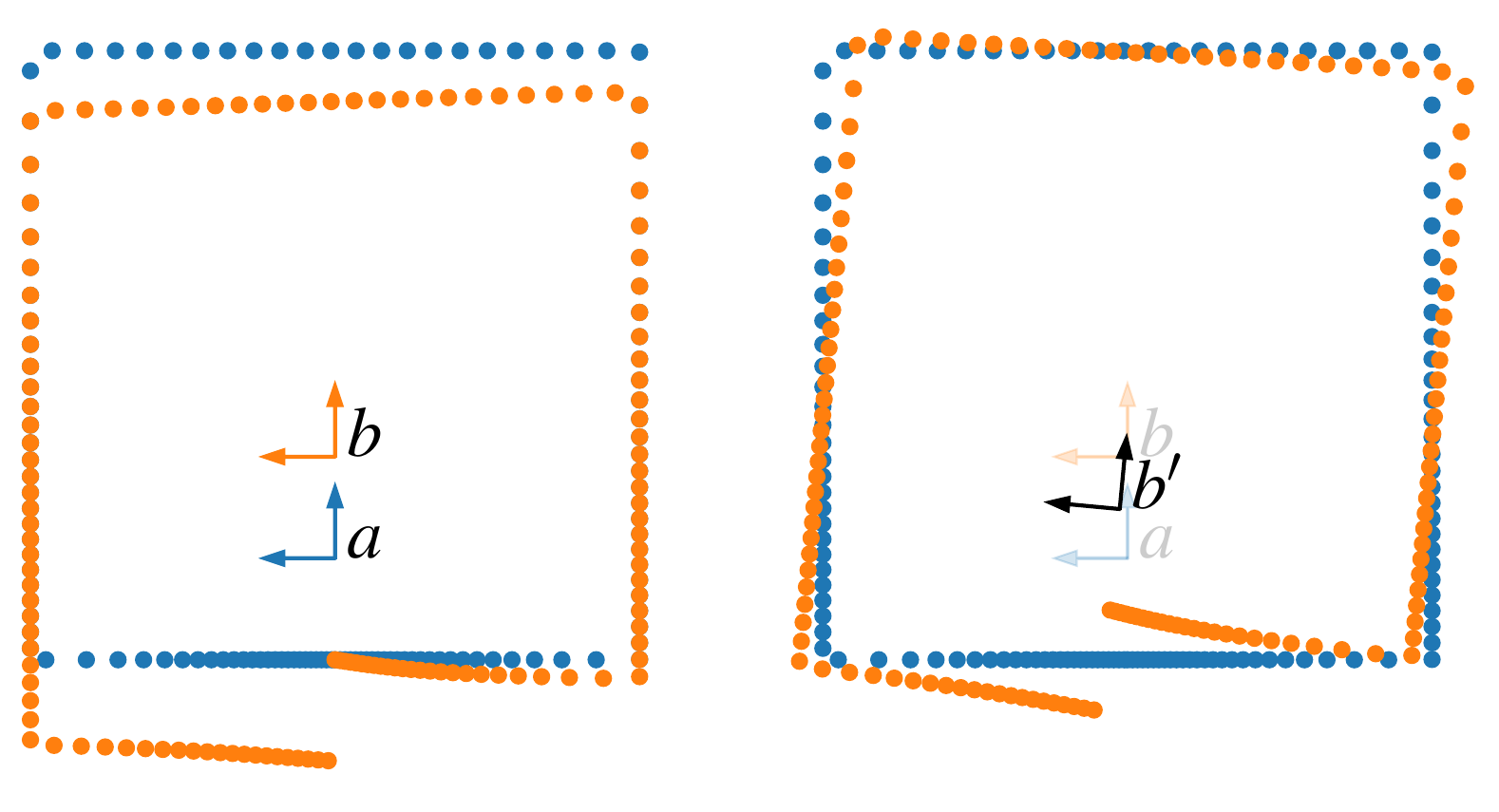}
    \caption{A toy example demonstrating the impact of motion on the registration of a square room. 
        Blue points represent the globally consistent map $\reference$ of the environment, with its origin pose $a$ marked with blue arrows. 
        Yellow points were acquired while the lidar was moving from pose $a$ to pose $b$, resulting in a heavily skewed point cloud.
        \emph{Left}: Before registration.
    \emph{Right}: After registration, where the new pose $b'$ is extracted from the resulting rigid transformation $\bm{T}$ computed by \ac{ICP}.
    Although the map combining both point clouds seems crisper on the right, the estimated pose of the lidar is wrong. 
    }
    \label{fig:skewing_impact_registration}
\end{figure}

For example, given a globally-coherent map $\reference$ and a skewed reading $\reading$ of a square room, the distortion will prevent both point clouds from aligning perfectly.
\ac{ICP} will thus find a transformation that distributes this alignment error throughout the whole reading $\reading$, as shown in~\autoref{fig:skewing_impact_registration}.
Therefore, because of the skew in the reading $\reading$, the rigid transformation $\T{}{}$ will not represent the actual displacement of the lidar since the last registration, causing a loss of accuracy in localization.
Furthermore, if a skewed reading $\reading$ is merged into the map, the newly added points will blur the overall structure, further increasing the loss of accuracy of the pose estimation.
Given that the mapping and the localization processes alternate at each scan, the overall robustness of the pose estimation framework can be rapidly jeopardized.
\vspace{-0.1in}

\subsection{De-skewing}
De-skewing methods take into account the lidar displacement to move each point $\point{i}$ in a common coordinate frame to correct a point cloud $\scan$ with a cardinality $\cardinality = |\scan|$.
One can decide that the origin of this coordinate frame is the pose of the lidar at the acquisition time of the first range measurement.
An assumption is made that each individual point $\point{i}$ has its own timestamp $t_i$.
Let $\T{i}{1}$ be the rigid transformation to convert between the lidar coordinate frame at time $t_i$ and the lidar coordinate frame at the time of acquiring the first point of the scan and $\point[i]{j}$ be the $j^{th}$ point of a scan expressed in the lidar coordinate frame at time $t_i$.
To construct the de-skewed point cloud $\hat{\scan}$, we need to compute
\begin{equation}
    \label{eq:deskewing}
	\hat{\scan} = \set{ \point[1]{1}, \T{2}{1} \point[2]{2}, \T{3}{1} \point[3]{3}, \;\cdots, \T{\cardinality}{1} \point[\cardinality]{\cardinality}} .
\end{equation}
In the remainder of this article, we will simplify the notation by stating that a point is expressed in its own coordinate frame when no subscript is present (i.e., $\point{i} = \point[i]{i}$).
From \autoref{eq:deskewing}, we can see that it is necessary to know the pose of the lidar each time a range measurement occurs in order to de-skew point clouds.
In other words, we need a finer-grained approximation of the trajectory over time than what a registration algorithm can give.

\subsection{Trajectory estimation}
\label{sec:odometry}
\begin{figure}[htbp]
    \centering
    \includegraphics[trim=0 0 0 10, width=0.8\columnwidth]{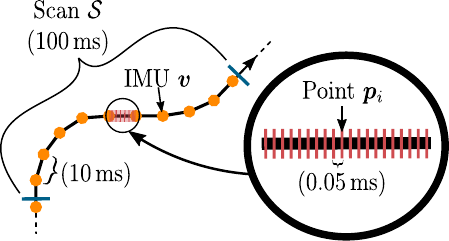}
    \caption{
    Measurement delays for all sensors used in this framework.
    The blue lines represent the start and end of a lidar scan, acquired at a rate of \SI{10}{Hz}.
    The orange dots represent the \ac{IMU} measurements, acquired at a rate of \SI{100}{Hz}, used to estimate the lidar velocity $\estlinvel$ during a scan.
    The red lines represent individual lidar measurements $\point{i}$, acquired at a rate of \SI{20000}{Hz}.
    }
    \label{fig:scan_rates}
\end{figure}

In order to de-skew point clouds in real-time, it is typical to use \ac{IMU} measurements and point cloud registration jointly~\cite{Zhao2019,Qin2020}.
Indeed, to compute the de-skewed point cloud $\hat{\scan}$, we must first compute the pose $\T{i}{1}$ of the lidar at each measurement time $t_i$, which occurs at a frequency of \SI{20000}{Hz}.
To do so, we use linear interpolation between lidar positions and orientations estimated at a frequency of \SI{100}{Hz}.
For lidar orientation estimation, a Madgwick filter \cite{Madgwick2011} based on gyroscope and accelerometer measurements is used.
It provides an orientation estimate at the same frequency as \ac{IMU} measurements (i.e., \SI{100}{Hz}).
As for the position of the lidar, it is estimated by integrating its linear velocity.
The latter is obtained by fusing two velocity estimates.
The first one corresponds to the single integration of the accelerometer measurements and yields estimates at the same frequency as \ac{IMU} measurements.
The other, which is more precise but less frequent, is computed by deriving the two last positions outputted by our localization and mapping framework and yields estimates at the scanning frequency (i.e., \SI{10}{Hz}).
With those two computed linear velocities, we obtain a reliable velocity estimate $\estlinvel$ at the same frequency as we receive \ac{IMU} measurements, without long-term drift.
This reliable estimate can then be integrated to compute the lidar position at a frequency of \SI{100}{Hz}.
Using the aforementioned techniques, we are able to compute a reasonable estimate of the pose of the lidar $\T{i}{1}$ within a scan.
The de-skewed point cloud $\hat{\scan}$ is then computed using~\autoref{eq:deskewing} and registered in the map using our localization and mapping framework to obtain a more precise pose estimate.
The frequencies at which the sensors operate and at which the different estimates are computed are shown in \autoref{fig:scan_rates}.

Although the reconstruction of a trajectory using an \ac{IMU} and a lidar yields a reasonable estimate, limitations remain.
As described by \citet{Underwood2010}, common error sources in mapping include sensor noise, timing errors, and sensor miscalibration.
Even though the trajectory is estimated over a short period of time (i.e., between two scans), accelerations need to be integrated twice, while the angular velocities need to be integrated once.
These integrals will accumulate error over the duration of the scan leading to an increase of uncertainty over the trajectory.
Smaller \ac{IMU} typically found in mobile robots rely on \ac{MEMS} to measure linear acceleration and rotational velocity.
Their measurements suffer from scaling errors and bias, component misalignment errors, non-linearities and vary depending on the temperature \cite{Qureshi2017}.
Both the \ac{IMU} and lidar must produce a globally coherent timestamp, which may require dedicated hardware to avoid drift and resolve any offsets.
Finally, to be able to estimate the trajectory tracking the centre of the lidar, the pose of the \ac{IMU} must be calibrated with respect the lidar.
This calibration will be estimated up to some accuracy, and will be subject to wear of the fixation during the use of the robot.
Any of these sources of uncertainty will produce a point cloud deformed with residual errors.
Thus, we propose a way to estimate such uncertainties and to exploit this information in the registration process.

\subsection{Registration point weighting models}
\label{sec:point_weight}

To limit the impact of uncertainty in the trajectory estimate, we propose to lower the weights $w_i$ of individual points $\point{i}$ of the reading $\reading$ and reference $\reference$ point clouds that were more probably affected by motion distortion.
The \ac{ICP} algorithm relies on the distance $\bm{e}$ between all matched points $m$ to minimize a cost function $\mathrm{J}$.
In this process, it is typical to compute a covariance matrix $\bm{\Sigma}_l$ representing the local structure around a point in the reference $\reference$.
In this work, we also add the covariance matrix $\bm{\Sigma}_n$ defining the sensor noise and express it as an isotropic noise using the pessimistic model of \citet{Pomerleau2012}, by using a scalar matrix with its diagonal elements as $\sigma^2_n$. 
Finally, related to the uncertainty on the position of a point $\point{i}$ caused by an error in the trajectory estimation, we added the covariance matrix $\bm{\Sigma}_s$.
Again, we make the assumption that the uncertainty defined by $\bm{\Sigma}_s$ is isotropic, by using a scalar matrix with its diagonal elements as $\sigma^2_s$.
Following the work of \citet{Babin2019a}, we can approximate the point-to-Gaussian cost function $\mathrm{J}_\text{p-g}$ of \ac{ICP}, defined as
\begin{equation}
\mathrm{J}_\text{p-g} =  \sum_{m=1} \left(\bm{e}^T \left(\bm{\Sigma}_l + \bm{\Sigma}_n + \bm{\Sigma}_s \right)^{-1} \bm{e}\right)_{m},
\end{equation}
to produce the more commonly used weighted point-to-plane cost function $\mathrm{J}_\text{p-n}$ using
\begin{equation}
\begin{split}
    \mathrm{J}_\text{p-g} &\approx \mathrm{J}_\text{p-n} \\
     &= \sum_{m=1} \left(w (\bm{e} \cdot \normal)^2 \right)_{m},
\end{split}
\end{equation}
where
\begin{equation}
    w = \frac{1}{\sigma^2_n + \sigma^2_s}
\end{equation}
and where the approximation is that the covariance $\bm{\Sigma}_l$, related to the local structure, is simplified to use only the surface normal vector $\normal$, and $\bm{e}$ is the euclidean distance between two matched points.
In the following sections, three different models to compute the skew-related position uncertainty $\sigma^2_s$ are proposed in increasing order of complexity.


\subsubsection{\ac{TW}}
The~\ac{TW} model only takes into account the time at which points were acquired since the beginning of the scan to compute their skew-related uncertainty.
The intuition behind this model is that, as more time passes since the beginning of the scan, the lidar can cover a greater linear and angular distance, thus point position uncertainty increases.
The position uncertainty $\sigma_{si}$ of the $i^{th}$ point of a scan is computed as $\sigma_{si} = c_1 t_i$. where $c_1 =0.25$ is a scaling factor identified empirically to account for modelling errors.

\begin{figure*}[htbp]
	\centering
	\includegraphics[width=\textwidth]{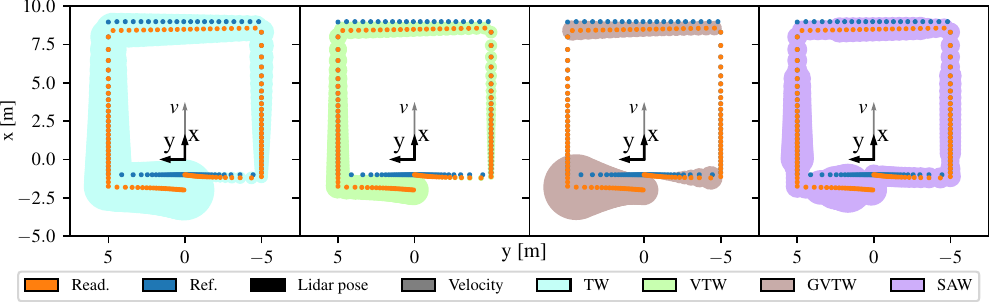}
	\caption{
	Uncertainty propagation on a toy example.
	The lidar scans anti-clockwise, starting from the negative $x$ direction.
	The measurement $\mathcal{P}$ is orange, and the reference $\mathcal{Q}$ is blue. 
	The initial lidar position is shown in black and its velocity direction in gray. 
	The shaded areas show the skewing uncertainty for each point of the measurement $\mathcal{P}$, for each model.
	}
	\label{fig:models_uncertainty}
\end{figure*}

The uncertainty grows linearly with time to be pessimistic and to account for possible bias in the velocities estimated by our odometry.
The left-most plot of~\autoref{fig:models_uncertainty} shows the uncertainty that would be computed for each point of the toy example measured point cloud by using the \ac{TW} model.

\subsubsection{\ac{VTW}}
\label{sec:model-1}
The \ac{VTW} model takes into account the lidar linear and angular speed uncertainties $\unlinspeed$ and $\unangspeed$ about each axis to compute the skew-related uncertainties of the points of a scan $\scan$.
The intuition behind this model is that if the estimated lidar linear and angular velocities $\estlinvel$ and $\estangvel$ are erroneous, point clouds will be inaccurately de-skewed, thus point position uncertainty will increase.
In order to compute the position uncertainty of a point, the point is rotated around the origin and translated by an amount proportional to $\unlinspeed$ and $\unangspeed$.
The translation parameters $\transparams = \begin{bmatrix}\transparam{x}, \transparam{y}, \transparam{z}\end{bmatrix}^T$ to apply to the points are computed in the following way:
\begin{equation}
	\label{eq:translation}
	\begin{split}
		\transparam{x} = \int_{t=0}^{t=t_i} \unlinspeed(\estlinspeed[x](t)) dt\text{,}
	\end{split}
\end{equation}
where $\estlinspeed[x](t)$ is the linear speed of the lidar along the $x$ axis at time $t$.
$\transparam{y}$ and $\transparam{z}$ are computed similarly, but along their respective axis.
The rotation parameters $\rotparams = \begin{bmatrix}\rotparam{x}, \rotparam{y}, \rotparam{z}\end{bmatrix}^T$ to apply to the points are computed in the following way:
\begin{equation}
	\label{eq:rotation}
	\rotparam{x} = \int_{t=0}^{t=t_i} \unangspeed(\estangspeed[x](t)) dt\text{,}
\end{equation}
where $\estangspeed[x](t)$ is the estimated angular speed of the lidar around the $x$ axis at time $t$.
$\rotparam{y}$ and $\rotparam{z}$ are computed similarly, but around their respective axis.
Our odometry linear $\unlinspeed(\cdot)$ and angular $\unangspeed(\cdot)$ speed uncertainty models are described in \autoref{sec:result-speed-noise}.
The position uncertainty $\sigma_{si}$ associated with the point $\point{i}$ of the currently-processed scan is computed as follows with this model:
\begin{equation}
	\label{eq:model-1}
	\sigma_{si} = \frac{c_2}{2} \max(\lVert (\rot{\point{i}, \pm \rotparams} \pm \transparams) - (\rot{\point{i}, \pm \rotparams} \pm \transparams) \rVert)\text{,}
\end{equation}
where $c_2 = 2$ is a scaling factor identified empirically to account for modelling errors and $\rot{\point{}, \rotparams}$ is the function that applies the rotation parametrized by $\rotparams$ to the point $\point{}$.
\autoref{eq:model-1} allows the computation of the highest de-skewing error a point can suffer from by testing 16 directions in which the velocity estimation error could be made.
The second plot from the left of~\autoref{fig:models_uncertainty} shows the uncertainty that would be computed for each point of the toy example measured point cloud by using the \ac{VTW} model.

\subsubsection{\ac{GVTW}}
The \ac{GVTW} model takes into account the lidar linear and angular speed uncertainties $\unlinspeed$ and $\unangspeed$ about each axis and the geometry of the environment to compute the skew-related uncertainties of the points of a scan $\scan$.
The intuition behind the \ac{GVTW} model is that erroneous velocity estimates $\estlinvel$ and $\estangvel$ will lead to an inaccurate de-skewing of a scan $\scan$, but not for points on all surfaces.
Indeed, if a surface is parallel to the direction in which a velocity estimate error is made, points measured on that surface will still be valid after de-skewing, since such an error will cause points to \emph{slide} along the surface.
To compute the position uncertainty of a point $\point{i}$, the point is rotated around the origin and translated by an amount proportional to to $\unlinspeed$ and $\unangspeed$.
Then, the distance that would have been measured if the laser had been fired in this direction is computed.
The translation $\transparams$ and rotation $\rotparams$ parameters to apply to the point are computed using \autoref{eq:translation} and \autoref{eq:rotation}.
The position uncertainty $\sigma_{si}$ associated with the point $\point{i}$ of the currently-processed scan is computed as follows with this model:
\begin{equation}
	\label{eq:model-2}
	\sigma_{si} = \frac{c_3}{2} \max( \lvert \distance{\point{i}, \pm \rotparams, \pm \transparams, \normal[i]} - \distance{\point{i}, \pm \rotparams, \pm \transparams, \normal[i]} \rvert)\text{,}
\end{equation}
with
\begin{equation}
    \label{eq:distance-function}
	\distance{\point{}, \rotparams, \transparams, \normal} = \frac{(\point{} - \rot{\transparams, \rotparams)} \bm \cdot \normal}{\rot{\frac{\point{}}{\lVert \point{} \rVert}, \rotparams} \bm \cdot \normal}\text{,}
\end{equation}
where $c_3 = 4$ is a scaling factor identified empirically to account for modelling errors, and $\normal[i]$ is the surface normal vector of $\point{i}$.
\autoref{eq:model-2} allows the computation of the highest de-skewing error a point can suffer from by testing 16 directions in which the velocity estimation error could be made.
\autoref{eq:distance-function} allows the computation of the distance that would have been measured for a point $\point{}$ if the laser had been fired from a different location and in a different direction.
We observed better results when points with an uncertainty above \SI{1}{m} were removed from the map.
This helps registration by matching scan points only with significant points of the map.
The third plot from the left of~\autoref{fig:models_uncertainty} shows the uncertainty that would be computed for  each  point  of  the  toy  example measured point cloud by using the \ac{GVTW} model.

\subsubsection{\ac{SAW}}
In order to evaluate the weighting models proposed previously, we use the \ac{SAW} model proposed by \citet{Al-Nuaimi2016} as a baseline because, to our knowledge, it is the only prior work addressing point cloud skewing using a point weighting model.
The \ac{SAW} model takes into account the horizontal scanning angle and surface curvature to compute the weight of each point in a scan.
Let $w_{si} = \cos(\scanangle[i] / 4)$ be the part of the weight of point $\point{i}$ which is associated with its horizontal scanning angle $\scanangle[i]$.
Let $w_{ci} = \max(0.25, \min(c_i/c_r, 1))$ be the part of the weight of point $\point{i}$ which is associated with its surface curvature $c_i$ and
where $c_r$ is an environment-dependent constant equal to the curvature of a highly curved surface.
The final weight $w_i$ of point $\point{i}$ is the maximum between $w_{si}$ and $w_{ci}$.

In the work of \citet{Al-Nuaimi2016}, this weighting function was used for the evaluation with non-simulated data.
The right-most plot of~\autoref{fig:models_uncertainty} shows the uncertainty that would be computed for each point of the toy example measured point cloud by using the \ac{SAW} model.

\section{Experimental Setup}
\label{sec:exp_setup}

\begin{figure}[htbp]
	\centering
	\includegraphics[width=0.95\columnwidth]{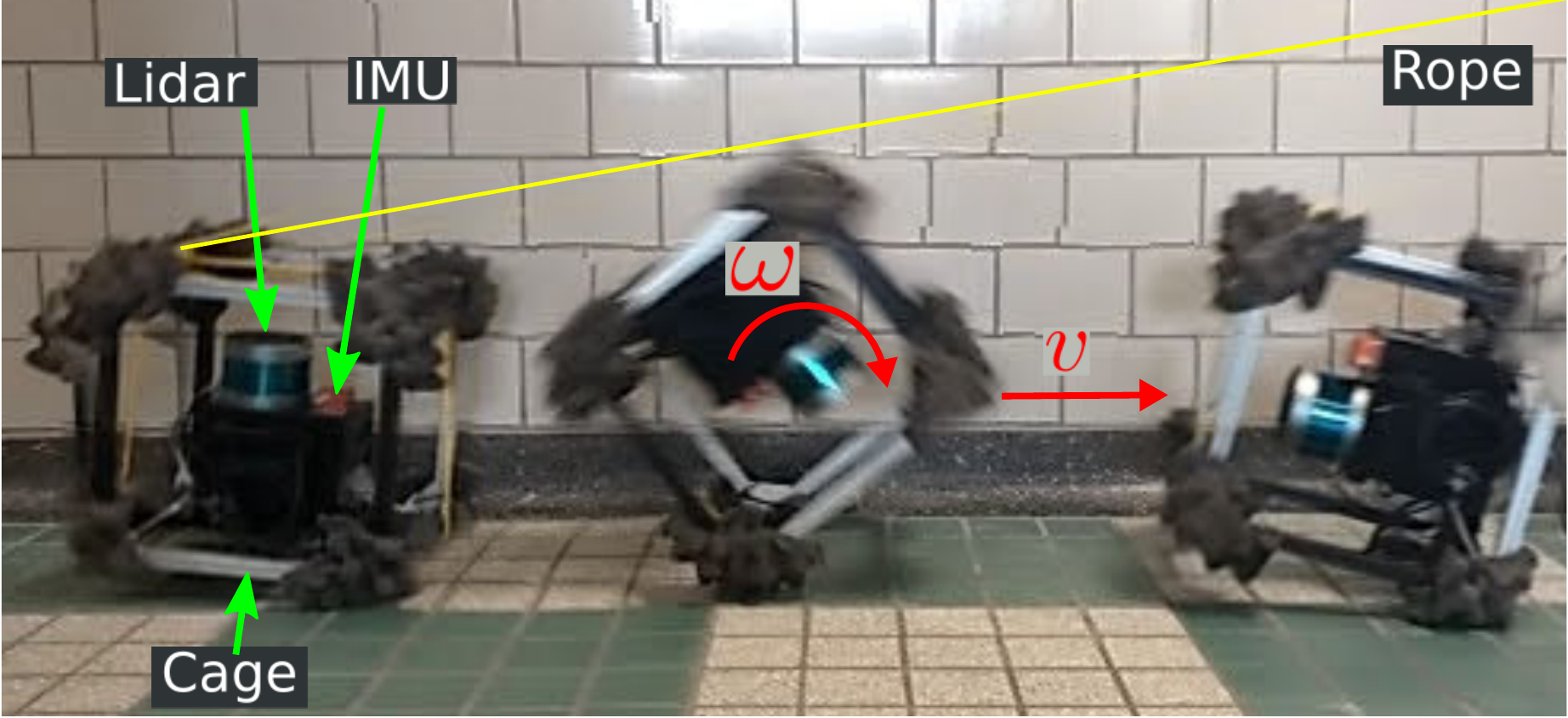}
	\caption{
	The experimental platform used for this work.
	A lidar and an \ac{IMU} are mounted on a sensing platform, which is protected by a cage. 
	A rope is attached to the platform and pulled on by an operator to induce high velocities and accelerations during lidar scans.
	}
	\label{fig:exp-setup}
\end{figure}

We conducted experiments with a Clearpath Husky on rugged terrain, as shown in~\autoref{fig:husky_ninja}, but the motions the robot underwent were not aggressive enough to induce visible scan deformations.
Therefore, to compare the models described in~\autoref{sec:point_weight}, we built the mobile sensing platform shown in~\autoref{fig:exp-setup}. 
A Robosense RS-16 lidar is located on top of the platform, producing scans at a rate of \SI{10}{Hz}.
Additionally, an Xsens MTI-30 \ac{IMU} returns its own body linear acceleration and angular velocity at a rate of \SI{100}{Hz}.
This platform was mounted on a \SI{30}{m} linear rail specialized for sensor calibration.
It was moved for a wide variety of linear velocities and accelerations, in both directions, while recording sensor measurements.
This same sensing platform was then mounted on a turntable and rotated in both directions while recording sensor measurements.
Both of the aforementioned experiments were conducted in order to submit the sensors to a maximum excitement range for linear and angular speeds. 
Afterwards, we added a protective cage to the platform, which can be seen in~\autoref{fig:exp-setup}. 
A rope was attached to the cage and an operator pulled on the rope in order to initiate the rolling of the cage on the ground at high angular velocities. 

\section{Results}
\label{sec:results}

Some of the weighting models described in \autoref{sec:point_weight} rely on having an estimate of the speed estimation uncertainties $\unlinspeed$ and $\unangspeed$ for each axis.
To establish the latter, we first modelled the linear and angular speed uncertainties for one axis of our odometry system from the collected data presented in \autoref{sec:result-speed-noise}.
We then compared the different weighting models on real lidar scans acquired at high linear and angular velocities and under high accelerations.
Finally, the effects on localization error and mapping are reported in \autoref{sec:result-cube}.

\subsection{Linear and angular speed uncertainties $\unlinspeed$ and $\unangspeed$}

\label{sec:result-speed-noise}
To compute the weights of the different models introduced in \autoref{sec:point_weight}, we must find a continuous function that allows the estimation of the linear speed uncertainty $\unlinspeed$ along an axis as a function of the estimated linear speed $\estlinspeed$ along that same axis.
Similarly, we should be able to compute the angular speed uncertainty $\unangspeed$ as a function of the estimated angular speed $\estangspeed$ around an axis.

To do so, we develop equations linking the uncertainties $\unlinspeed$ and $\unangspeed$ to registration residuals that we can measure in our data together with $\estlinspeed$ and $\estangspeed$.
De-skewing a point cloud with a linear velocity error $\epsilon$ is equivalent to not applying de-skewing to a point cloud acquired with a lidar travelling at a velocity of $-\epsilon$.
Therefore, the position error of a point $\point{i}$ along an axis can be retrieved by multiplying its timestamp $t_i$ with the linear speed estimation error $\unlinspeed$ along that axis.
Assuming that the velocity throughout a scan is constant, 
the mean residual registration error $r_v$ caused by de-skewing a point cloud using a linear speed estimate with an error $\unlinspeed$ along an axis can be computed.
Indeed, it is approximately the displacement of the lidar in the middle of its scan.
Thus, we have
\begin{equation}
    \label{eq:linear-speed-noise}
    r_v \approx \frac{\unlinspeed  \tau}{2}\text{,}
\end{equation}
where $\tau$ is the time needed to complete a whole scan.
This equation allows us to express our linear speed uncertainty model $\unlinspeed$ along an axis as a function of the linear speed $\estlinspeed$ estimated by our odometry along that axis.
Similarly, it can be shown that the mean residual registration error norm $r_\omega$ can be computed as a function of the angular speed estimation error $\unangspeed$ around an axis using the following equation:
\begin{equation}
    \label{eq:angular-speed-noise}
    r_{\omega} \approx \frac{\unangspeed \tau d}{2}\text{,}
\end{equation}
where $d$ is the mean distance measurement of the points in the point cloud.

In order to obtain the uncertainty models $\unlinspeed$ and $\unangspeed$ from real data, we mounted the platform on the rail and the turntable described in \autoref{sec:exp_setup}, to acquire scans at different linear and angular speeds.
Then, we computed the mean norm $r_v$ or $r_\omega$ of registration residuals of the de-skewed scans against a previously-built dense 3D map of the environment.

\begin{figure}[htbp]
    \centering
    \includegraphics[trim=0 0 0 0, width=\columnwidth]{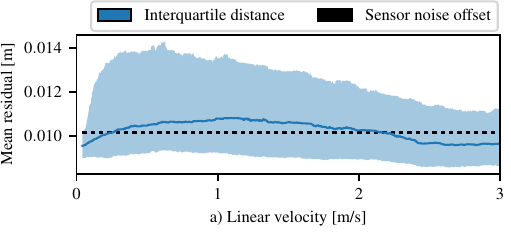}
    
    \vspace{.05in}

    \includegraphics[width=\columnwidth]{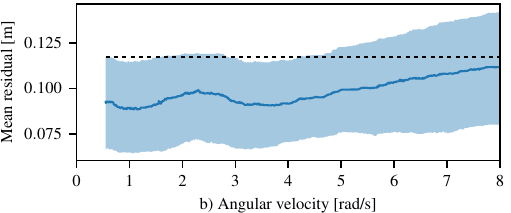}
    \caption{
    Registration residual as a function of a) linear speed and b) angular speed. The median is the solid blue line, and the shaded zones represent the first and third quartiles.  
    The black dotted line represents the residual offset caused by sensor measurement noise.
    For a), only residuals from surfaces \emph{perpendicular} to the lidar motion are used.
    }
    \label{fig:modelization}
\end{figure}

For the linear speed uncertainty $\unlinspeed$, these residuals were computed using the euclidean distances $\bm{e}_i$ projected on surface normals $\normal[i]$ only for points $\point{i}$ lying on surfaces perpendicular to lidar motion. 
Those points were selected because, unlike points lying on surfaces parallel to the motion, their error increases when inaccurate de-skewing caused by linear motion occurs.
Since the map is assumed to be accurate, the scan registration residuals were therefore due to \emph{i)} the noise in lidar measurements and \emph{ii)} the skew caused by inaccurate de-skewing.
\autoref{fig:modelization}-a shows the median and quartiles of the mean registration residuals as a function of the linear speed at which point clouds were acquired. 
If the lidar is not moving, the velocity estimation error made by our odometry system should be low enough to consider that all registration residuals come from lidar measurement noise.
Furthermore, the registration residuals caused by measurement noise are assumed to be independent of lidar velocity.
Therefore, by removing the registration residual computed at a speed of \SI{0}{m/s} from computed residuals $r_v$, we obtain the residual caused by inaccurate de-skewing.
This error is depicted in \autoref{fig:modelization}-a with a black dotted line.

After converting the residuals shown in \autoref{fig:modelization}\mbox{-a} to linear speed uncertainties using \autoref{eq:linear-speed-noise}, the third quartile of the uncertainty distribution is fitted using a log-normal function.
This provides an empirical model of the linear speed uncertainty $\unlinspeed$ along an axis of our odometry system as a function of the linear speed $\estlinspeed$ it estimates along that axis, such that
\begin{equation}
    \unlinspeed(\estlinspeed) = \frac{\linscale}{\shape \estlinspeed \sqrt{2 \pi}} \exp\left(-{\frac{\log^2(\frac{\estlinspeed}{\scale})}{2\shape^2}}\right),
\end{equation}
where $\shape=1.1$ is the shape factor, $\scale=1.9$ is the scale factor and $\linscale=0.222$ is the function gain.
These values were empirically identified based on the data presented in \autoref{fig:modelization}-a.

Similarly, for the angular uncertainty model $\unangspeed$, we acquired scans at different angular speeds and accelerations and computed the mean residual registration error norms $r_\omega$ using a dense 3D map of the environment.
However, to highlight scan distortions, the euclidean distance $\bm{e}_i$ of all points $\point{i}$ in the map was used.
Euclidean distance was used because we observed that it increases more than its projection on surface normals when subject to inaccurate de-skewing caused by rotational motion.
\autoref{fig:modelization}-b shows the median and quartiles of registration residuals as a function of the angular speeds at which point clouds were acquired.
After converting the residuals shown in \autoref{fig:modelization}-b to angular speed uncertainties using \autoref{eq:angular-speed-noise}, the third quartile of the uncertainty distribution is fitted using a polynomial function.
This provides an empirical model $\unangspeed(\estangspeed) = (\estangspeed / \angscale)^3$
of the angular speed uncertainty $\unangspeed$ around an axis of our odometry system with a scaling parameter $\angscale$.
We have empirically identified $\angscale=16$.
Although our odometry system is quite generic, the aforementioned rail and turntable experiments can be repeated with different systems to identify better-suited values for $\shape$, $\scale$, $\linscale$ and $\angscale$.

\subsection{Impact of proposed method on localization and mapping}
\label{sec:result-cube}

Using the platform and protective cage described in \autoref{sec:exp_setup}, we acquired scans at linear speeds and accelerations up to \SI{3.5}{m/s} and \SI{200}{m/s^2} and angular speeds and accelerations up to \SI{11}{rad/s} and \SI{800}{rad/s^2} on a total of 46 runs.
For each run, we recorded scans and used the \ac{ICP} algorithm with the methods described in~\autoref{sec:point_weight} to compute the transformation between the initial and final poses of the lidar.
No prior map of the environment was used to compute these transformations.
They were computed using only the measurements acquired by the platform during a run.
For all runs, we used point-to-plane minimization combined with \ac{FRMSD} \cite{Phillips2007} to handle varying overlap ratios between point clouds.
We also used a voxel grid with a cell size of \SI{5}{cm} to down-sample input point clouds.

\begin{figure}[htbp]
    \centering
    \includegraphics[trim = 0 0 0 2, width=\columnwidth]{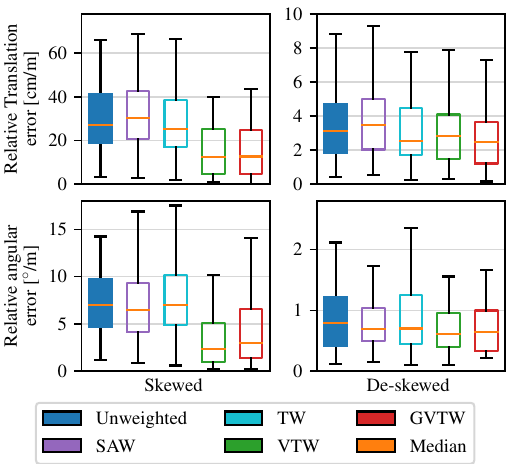}
    \caption{
    Relative translation and angular errors for all models.
    Results were gathered offline, based on 46 runs which consisted of rolling a platform  with a lidar and an \ac{IMU}  attached to it. 
    The error is computed as the difference between the final estimated pose by each model and the ground truth pose. 
    The ground truth is computed by registering the final recorded point cloud in a static map.
    }
    \label{fig:result-models}
\end{figure}

To compute the ground truth transformations between the initial and final poses of the lidar, we registered the first and last scans of each run in a dense 3D map of the environment built while moving slowly.
This allowed us to compare the impact of de-skewing and point weighting on localization and mapping accuracy, as shown in \autoref{fig:result-models}.
The translation and rotation localization errors of the different models are relative to the distance travelled by the platform in a run.
In the graph, the blue boxes are the resulting localization errors when no point weighting is used.
The left plots show the results of using point weighting models without de-skewing, and the right plots show the result of their use after point cloud de-skewing. 
As can be seen, without any point weighting model and without de-skewing, the median of the relative localization error is \SI{27.23}{cm/m} and \SI{7.04}{deg/m}.
However, when point cloud de-skewing alone is used, the median translation error decreases to \SI{3.11}{cm/m} and the median rotation error decreases to \SI{0.79}{deg/m}.
These results show that de-skewing methods can significantly enhance localization and mapping accuracy.
Indeed, by correcting the distortion in point clouds, de-skewing methods allow greater overlaps between the scans and the map, leading to a crisper map, and thus to better localization.
However, if no de-skewing method can be applied, the \ac{VTW} point weighting model allows a decrease in the median localization error by a factor 2.2 in translation and by a factor 3.0 in rotation, which is the best improvement out of all of the studied models.
However, the best combination consists of adding the \ac{GVTW} model after point cloud de-skewing, where the median localization error is reduced by \SI{9.26}{\%} in translation and by \SI{21.84}{\%} in rotation when compared to not using point weighting on a de-skewed point cloud.
Both the \ac{VTW} and \ac{GVTW} models also allow the interquartile range to be significantly reduced, meaning that they produce a more precise localization.
It is also worth noting that the \ac{SAW} model~\cite{Al-Nuaimi2016} does not improve the mean localization accuracy in our tests.

\begin{figure}[htbp]
    \centering
    \includegraphics[trim = 0 0 0 0, width=\columnwidth]{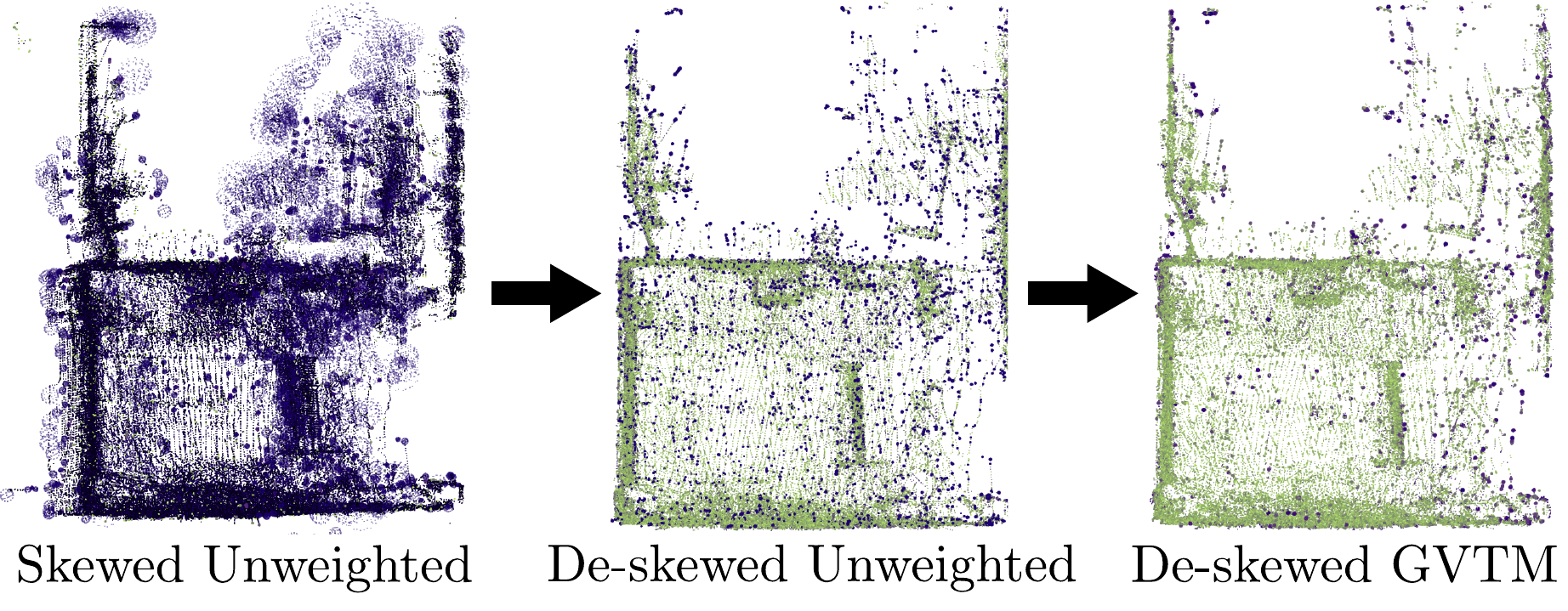}
    \caption{
    Top view of an indoor garage mapped under high velocities and accelerations.
    Points in dark purple have high error compared to a ground truth map.
    }
    \label{fig:final_maps}
\end{figure}

Demonstrating qualitatively the impact of our weights on mapping, \autoref{fig:final_maps} shows a 3D reconstruction of an environment built while applying peak accelerations up to \SI{194.98}{m/s^2} on our sensing platform.
All three maps were built using the same data, with the \ac{GVTW} weighting model only used for the rightmost map.
The error was computed against a ground truth map built while moving slowly the sensors.
De-skewing significantly impacts the mean error by reducing it from \SI{21.95}{\cm} to \SI{7.27}{\cm}.
Further improvement is gained by using \ac{GVTW}, leading to a mean error of \SI{6.86}{\cm}.
For example, one can observe an improvement in accuracy on the left wall.

\section{Conclusion}
\label{sec:conclusion}
In this paper, we address the challenge of localization accuracy when a mobile robot is experiencing extreme motions.
To this effect, we introduced three novel motion uncertainty-based point weighting models to increase the robustness of point cloud registration algorithms under such conditions.
We quantified that using a de-skewing solution reduces the localization error by an order of magnitude.
However, uncertainty remains in the trajectory estimation generating a localization error that can be further improved by using our solution.
Future improvements include removing the assumption that a scan is generated only after a full revolution, removing empirically-identified scaling factors from our method and testing with a hand-thrown device to expose our solution to even more extreme scenarios.


\section*{Acknowledgment}

This research was supported by the Natural Sciences and Engineering Research Council of Canada (NSERC) through the grant CRDPJ 527642-18 SNOW (Self-driving Navigation Optimized for Winter).


\printbibliography

\end{document}